\documentclass{cap2026}

\usepackage{fancyvrb}
\usepackage{amsfonts} 
\usepackage{amssymb}
\usepackage{amsmath}
\usepackage{booktabs} 
\usepackage{cleveref}
\usepackage{graphicx}
\usepackage{latexsym}
\usepackage{algorithm}
\usepackage{algpseudocode}
\usepackage{amsmath}
\usepackage{multirow}
\usepackage{makecell}
\usepackage{stfloats}
\usepackage{arydshln}
\usepackage{subcaption}

\title{Attention-based Experience Replay Framework for Continual Learning of Agnostic Time Series Forecasting Models}
\author[1]{Quentin Besnard}
\author[1]{Nicolas Ragot}
\affil[1]{University of Tours, LIFAT\thanks{name.surname@univ-tours.fr}}
\date{}

\begin{document}
\nolinenumbers
\maketitle

\begin{abstract}
Deep learning has led to remarkable progress in artificial intelligence, particularly in robotics, imaging and sound processing. However, a major limitation of neural networks remains their strong dependence on large and stationary datasets. In many real-world applications, these conditions are rarely met due to evolving and dynamic environments where data distributions change over time. Continual learning aims to address this challenge by developing models capable of adapting incrementally while maintaining a balance between stability and plasticity under computational constraints. In this work, we introduce a novel framework for continual time series forecasting, designed to extend existing static forecasting models commonly used in the literature by incorporating an Experience Replay strategy guided by Attention mechanisms. This approach allows the model to adapt dynamically to new contexts while preserving prior knowledge, effectively mitigating catastrophic forgetting. The framework is evaluated on standard forecasting benchmarks as well as on a piezometric dataset exhibiting diverse temporal behaviors. Results show that our approach effectively increases or maintains predictive performance over time while reducing retraining costs and data requirements, thus facilitating the deployment of forecasting models in dynamic and real-world settings.   
\medskip

\end{abstract}

\section{Introduction}
Recent advances in deep learning have led to significant progress in both short-term and long-term time series forecasting. These models are trained on historical data assuming the stationarity of the data over the time. However, in practice, the statistical characteristics of time series can change over time due to shifts in user behavior, economic factors or environmental conditions \cite{Besnard2024}. This phenomenon is commonly referred to as "data drift" or "concept drift" \cite{Gama2014}. Such changes can significantly impact the quality of predictions if the model is not updated.

To overcome this challenge, approaches based on periodic retraining have been widely studied, where the model is retrained from scratch at regular time intervals (\textit{eg}. joint training \cite{Chen2024}). However, these approaches require storing all previously observed data and their adaptations can be computationally expensive, which limits their applicability in real-world scenarios with constrained resources (such as data accessibility or latency issues). Moreover, over time, the drifted data tend to become negligible compared to the large volume of stored data, making effective adaptation increasingly difficult. 

Recently, continual learning (CL) \cite{Qu2021} has emerged as a promising approach for handling non-stationary environments by managing the trade-off between plasticity and stability. In the context of supervised learning, CL enable a model to learn incrementally across sequential tasks while avoiding the phenomenon of catastrophic forgetting, \textit{ie}. the loss of performance on previously learned tasks \cite{Matthias22}. Among various CL strategies, Experience Replay (ER) \cite{Bagus2021} has proven to be highly effective in maintaining past knowledge while improving model performance by reusing a subset of historical data when training on new data. However, contrary to the extensive work conducted on classification tasks, the application of CL to time series forecasting remains underexplored. 

This research focuses on improving Continual Learning with Experience Replay for multivariate Long Term Time Series Forecasting (LT-TSF) model adaptation. It explores different strategies for sampling and managing a memory buffer of limited size containing representative samples. We propose an attention-based sampling strategy and compare it with traditional sampling methods from the literature, such as random sampling \cite{Schillaci2021} and loss-based sampling \cite{He2021}. These experience replay strategies are evaluated against standard training approaches, including single-phase batch learning and continual retraining with accumulated data (joint training).

Our main contributions are as follows:
\begin{itemize}
    \item We formalize the adaptation of predictive models for time series as a continual learning problem.
    \item We propose a model-agnostic adaptation framework for predictive models.
    \item We study updating strategies and selection methods in the context of experience replay.
    \item We empirically evaluate our proposals on several real-world time series datasets with multivariate forecasting, such as ETT and Weather datasets (which are standard benchmarks) \cite{Gong2023}, as well as Piezometric data in the context of our research for a digital twins application.
\end{itemize}

\section{Motivation and Related Work}
\subsection{Time Series Forecasting}
Time Series Forecasting is a fundamental task that involves predicting the future values of a given time series based on its past observations. Many methods have been proposed to address this problem, ranging from classical models (such as ARIMA and Exponential Smoothing), to machine learning approaches like random forests, support vector machines (SVM) and more recently deep learning models.

In the field of deep learning, Recurrent Neural Networks (RNNs) and their variants, such as LSTM and GRU, have long been used to model temporal dependencies \cite{Hewamalage2021}. More recently, attention-based architectures such as Informer \cite{Zhou2020}, Autoformer \cite{Wu2021}, and FEDformer \cite{Zhou2022} have achieved strong forecasting performance by capturing long-term dependencies. However, linear-based approaches have challenged the superiority of Transformers \cite{Zeng2022}, despite newer Transformer models continuing to improve benchmark results \cite{Nie2022}, thereby maintaining a close debate between forecasting architectures.

Most of these models assume that time series data are stationary, meaning that their statistical properties do not change over time. However, this assumption is often incorrect in practice, where distributional shifts may occur, for example, due to changes in human behavior, shifts in seasonality or unforeseen events. Consequently, it is crucial to develop methods that allow for dynamic adaptation of models to new data. In addition, those advanced models typically require large amounts of data and extensive historical records, whereas more simpler models can sometimes achieve comparable results. In this context, it appears relevant to continuously train a reliable baseline model to strengthen its robustness and adaptability. 

\subsection{Continual Learning}
Continual Learning (CL) aims to train models on new data while preventing catastrophic forgetting. This concept refers to the loss of knowledge over time which produces an overall decline in performances. Different main strategies for continual learning have emerged in the literature and can be categorized as follows  \cite{Chen2024}:
\begin{itemize}
    \item Regularization: Prevents significant changes to the model’s weights that are considered crucial for previously learned tasks \cite{Aich2021}.
    \item Dynamic Architecture: Adds or modifies parts of the model during the learning process \cite{Ye2023}.
    \item Experience Replay: Stores a subset of examples from previous tasks and continuously reintroduces them during the learning of new tasks \cite{Buzzega2020}.
\end{itemize}

Among these approaches, experience replay (ER) has been proven to be particularly effective, easy to implement, and applicable across various learning scenarios. ER offers a more efficient alternative to traditional full retraining on all data collected so far and to fine-tuning, as both of these strategies tend to prioritize either stability or plasticity at the expense of the other. ER operates by storing previously observed data and sampling  past examples to be reused during training on new data. This mechanism helps maintain performance on earlier contexts while still allowing the model to adapt to recent changes. While ER in CL has been extensively studied in the context of classification tasks (\textit{eg}, images or text), its application to time series forecasting remains underexplored \cite{Gupta2022}. 

Although effective, ER remains limited in practice for several reasons: storage cost; computation to continuously scan the memory to select samples; requirement for constant access to past data (which can be problematic when dealing with sensitive information or for storage reasons) and tendency to overwhelm new data with an increasing volume of replayed samples in order to mitigate catastrophic forgetting. Random sampling strategies provide significant improvements over naive adaptation, but their performance often degrades due to the lack of specialization. 

In this work, we explore the use of a strictly constrained ER for continual learning time series forecasting models. Our approach relies on a fixed-size buffer (to mitigate increasing data storage) and on an efficient data management since forgotten samples cannot be recovered. To address this limitation, we introduce a dedicated attention-based module designed to analyze incoming data and to perform informed sampling to manage the buffer content effectively.

\section{Problem Statement and proposed methodology}
\subsection{Problem Statement}
\label{ProbStatement}
In this work, we address the problem of multivariate time series forecasting within a continual learning framework, where data arrives sequentially in the form of batches and the model must be continuously adapted to these new data. This problem raises several challenges:
\begin{itemize}
    \item Non-stationarity: the data distribution may change over time (trend changes, seasonality, breaks), requiring dynamic adaptation.
    \item Learning without access to all previous data: in many use cases (embedded sensors, industrial flows, sensitive data), it is not feasible to store and reuse all past data. Moreover the systematic storage of massive data, whatever their relevance, becomes an environmental problem that must be tackled.
    \item Preservation of past performance: the model must maintain its performance on previous contexts while adapting to new ones (limited catastrophic forgetting).
\end{itemize}

We propose to tackle these challenges by providing an ER CL framework that is model agnostic. This framework integrates a new experience replay approach based on an Attention Model for sampling and managing a fixed size buffer of examples to be used during CL.

More formally, we consider:
\begin{enumerate}
    \item a given forecasting model that uses a lookback window of size $input\_len$ and predict a set of values (targets) depending on horizon $pred\_len$;
    \item a stream of data batches denoted ${B_1, B_2, ..., B_T}$, where each batch $B_t$ consists of a set of multivariate time series sequences of length $input\_len$, accompanied by their target $pred\_len$ values.
\end{enumerate} 
At each step $t$, the goal is to update a predictive model $f$, parameterized by $\theta$, based on the current batch $B_t$, without forgetting the knowledge learned from previous batches $B_1, B_2, ..., B_{t-1}$. 

Note that it is crucial to account for temporal lag when computing the loss during adaptation, particularly in continual learning. While one could compute the loss by comparing predictions with future targets in the dataset, this does not reflect real-world conditions. At each step $t$, the model receives a sequence $x_{t-input\_len:t}$ and predicts  $\hat{y}_{t+1:t+pred\_len}$, where target values $y_{t+1:t+pred\_len}$ will only be observed at time $t+pred\_len$. To respect this causal constraint, loss evaluation $L_t$ must be deferred until the future targets are observed and updated at time $t$ based on input data observed before $t-pred\_len$. Consequently, any performance evaluation or calibration in continual learning must be retrospective, ensuring temporal consistency with real-world deployment. This delay limits the model’s ability to immediately correct predictions (loss of plasticity that depends on $pred\_len$).

\subsection{Proposed Framework overview}
The proposed framework allows transforming a forecasting model from conventional train-test  learning scenario into a continual learning version. It can work with different predictors, but we used here the PatchMixer model \cite{Gong2023} as the prediction head. Beside the prediction head we propose an adaptation module along with several strategies.

To ensure long-term sustainability, the model operates under deliberate resource constraints. A fixed-size buffer is used for experience replay and a relaxed online learning setup is enforced: once data leave the model or the buffer, they cannot be reused. These constraints emulate realistic industrial scenarios, such as digital twins \cite{Lombardo2024}, where data volumes are too large for systematic storage.

Continual model adaptation occurs at two complementary levels:
\begin{itemize}
    \item Direct adaptation on the most recent batch, maintaining alignment with real-world dynamics through single-epoch updates using standard backpropagation. For each batch $B_t$ the forecasting model is briefly fine-tuned on this newly data for a quick adjustment.
    \item Strategy-based adaptation via Experience Replay (ER, Algorithm~\ref{alg:attn_replay}), which mixes past and current data from the buffer to mitigate catastrophic forgetting.
\end{itemize}

The buffer is updated following a predefined selection strategy which could be:
\begin{itemize}
    \item Random sampling (classical, stratified, etc) \cite{Schillaci2021}.
    \item Selection based on statistical metrics (\textit{eg}. best/worst performance samples) \cite{Han2022}.
    \item Learning-based selection using auxiliary models (\textit{eg}. attention, autoencoders evaluating reconstruction quality, etc) \cite{He2021}.
\end{itemize}

These different approaches are typically closely tied to their specific use cases. In our experiments, we compared three strategies: random sampling, loss-based selection and our proposed attention-based learning module for data selection. Each strategy is controlled by three external parameters:
\begin{itemize}
    \item Buffer size ($BS$): maximum capacity for stored samples;
    \item Past replay ratio ($k^{past}$): proportion of past data used during replay;
    \item Current replay ratio ($k^{curr}$): proportion of recent data used.
\end{itemize}

For instance, setting both ratios to 0.2 means that only 20\% of buffer with past data and 20\% of new data from current batch are sampled and used. Thus, an efficient data selection strategy becomes critical.

The following subsections detail the predictive head used and our proposed attention-based selection mechanism for experience replay.

\subsection{Predictive head for multivariate Time Series Forecasting}
Our continual learning framework builds upon PatchMixer \cite{Gong2023}, a recent state-of-the-art model for multivariate time series forecasting. The implementation  relies on the PatchMixer GitHub repository\footnote{GitHub repository: "\href{https://github.com/Zeying-Gong/PatchMixer/tree/main}{PatchMixer}"}, which contains the necessary tools for conventional training and testing.

PatchMixer achieves competitive accuracy while being lighter and more efficient than attention-based architectures. Unlike Transformer models that rely on complex attention mechanisms, PatchMixer follows a fully convolutional design. It divides the input sequence into temporal patches (sub-sequences), which are successively encoded, mixed, and decoded through convolutional layers:
\begin{itemize}
\item Patch Encoding: The input series is segmented into fixed-length patches, each encoded via convolutions to extract local features.
\item Patch Mixing: Encoded representations are combined through additional convolutions, capturing inter-patch dependencies without attention.
\item Decoding and Projection: The aggregated features are projected to forecast future values.
\end{itemize}
In this work, PatchMixer serves as the backbone of our continual learning framework, supporting progressive and stable adaptation to non-stationary data streams.

\subsection{Attention-based Experience Replay}
We propose a continual adaptation strategy inspired by the ER paradigm in CL \cite{Kirkpatrick2017}. At each time step $t$, a quick adaptation of the predictor is applied on batch $B_t$, after which the full framework (forecasting and sampling modules) is updated on $M_t$, a memory buffer updated with $B_t$ and $M_{t-1}$ using ER sampling in accordance with the constraint defined in part \ref{ProbStatement}.

\begin{figure}[!ht]
\centerline{\includegraphics[width=0.8\textwidth]{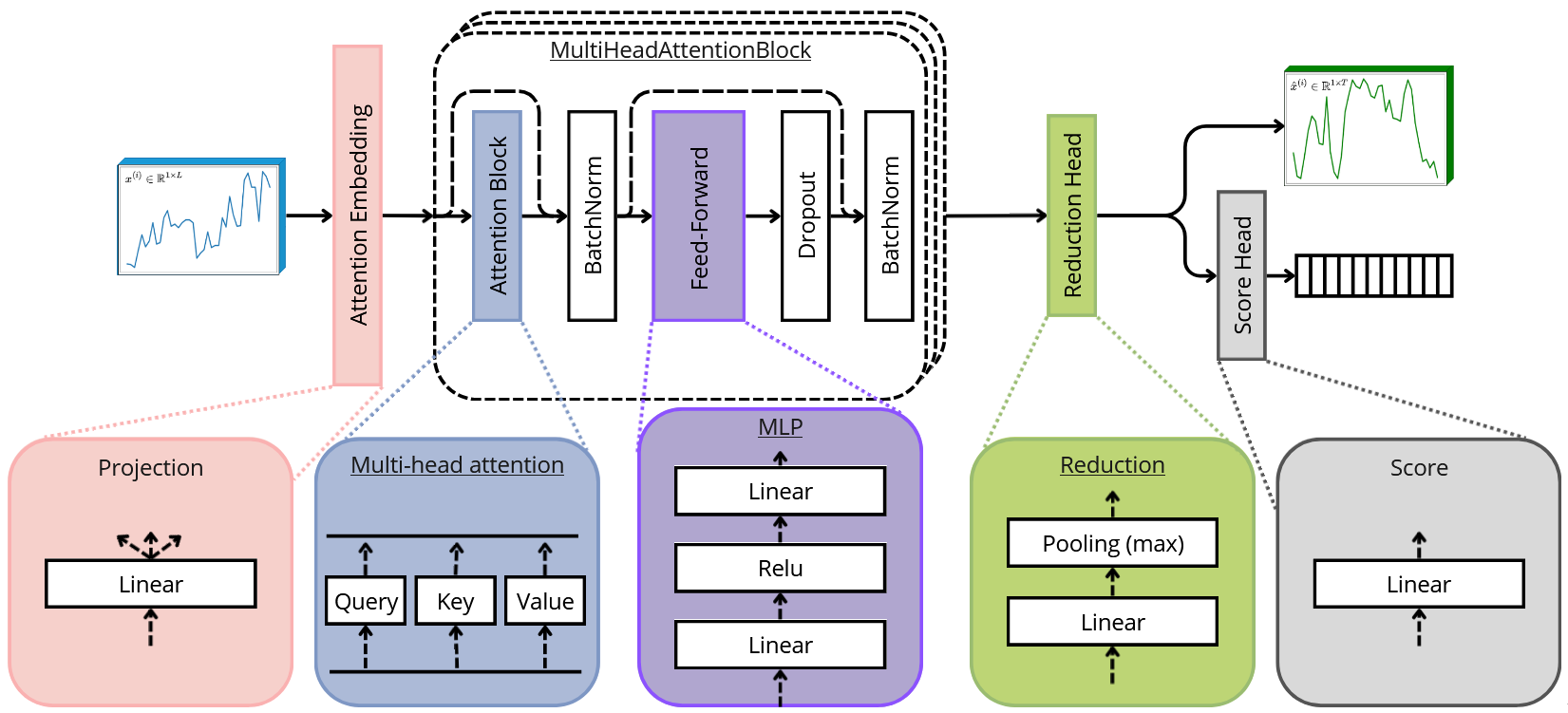}}
\caption{Sampling Multi-Head Attention Model (MHAM).}
\label{fig1}
\end{figure}

To manage this buffer and select only useful data to keep from previous buffer and current batch available, we propose an attention-driven sampling strategy, illustrated in Figure~\ref{fig1}. 
Prior work has shown that attention can enhance model adaptability in continual learning \cite{Sokar2021}. Unlike random or heuristic sampling, our Multi-Head Attention Module (MHAM) learns to identify the most informative samples through a relevance-scoring process. Composed of three Transformer blocks, MHAM captures intra-batch dependencies and assigns scores that guide memory updates, ensuring that the buffer retains the most valuable information for forecasting. MHAM is jointly learnt (continuously) with the predictive model, using forecasting loss as supervision. It is trained during a first training phase to learn key parameters and deployed seamlessly during a CL testing phase. The code can be found in the University of Tours LIFAT GitLab\footnote{LIFAT GitLab repository: "\href{https://scm.univ-tours.fr/projetspublics/lifat/junon-prediction.git}{CL-sampling}"}

\begin{algorithm}[!ht]
\caption{Experience Replay with Attention-based Sampling}
\label{alg:attn_replay}
\begin{algorithmic}[1]
\Require Predictive model  $f:\theta_1$; attention-based sampling model  $\mathcal{G}:\theta_2$; sequential data  $X_{t=1}^{T}$; learning rates  $\eta_1$ and $\eta_2$; sampling rates $k^{curr}$ and $k^{past}$
\Ensure Continuous update of $f$ and $\mathcal{G}$ with new incoming batches ($B_t$)
\State Initialize the memory buffer $\mathcal{M} \gets s^{curr}_j$ from scores $s^{curr}_i \gets \mathcal{G}(B_0)$
\For{each iteration $t$ from $t \gets 1$}
    \State Receive a new data batch $B_t$ at time step $t$
    
    \State Compute the scores $s^{curr}_i$ for each $x_i$ in $B_t$ using the sampling model, $s^{curr}_i \gets \mathcal{G}(B_t)$
    \State Compute the scores $s^{past}_i$ for each $x_i \in M_{t-1}$ using the sampling model, $s^{past}_i \gets \mathcal{G}(M_{t-1})$
    
    \State Select a subset $s^{curr}_j \subset s^{curr}_i$ based on the score distribution, $s^{curr}_j  \gets \text{Sample}(s^{curr}_i, k^{curr})$
    \State Select a subset $s^{past}_j \subset s^{past}_i$ based on the score distribution, $s^{past}_j  \gets \text{Sample}(s^{past}_i, k^{past})$
    
    \State Construct the memory buffer $\mathcal{M}_t \gets s^{curr}_j \cup s^{past}_j$
    
    \State Update the model parameters $f$: $\theta_1 \gets \theta_1 - \eta_1 \nabla_{\theta_1} \mathcal{L}_t(\mathcal{M}_t, \theta_1)$
    \State Update the model parameters $\mathcal{G}$: $\theta_2 \gets \theta_2 - \eta_2 \nabla_{\theta_2} \mathcal{L}_t(\mathcal{M}_t, \theta_2)$
\EndFor
\end{algorithmic}
\end{algorithm}

The complete Attention-based ER procedure is described in Algorithm~\ref{alg:attn_replay}.
At each iteration:
\begin{itemize}
\item MHAM computes relevance scores to all samples in $B_t$ and $M_{t-1}$. First, we initilize the buffer $M$ with $B_0$.
\item Based on scores and sampling rates ($k^{past}$, $k^{curr}$), two subsets $S$ are selected and merged to form the updated buffer $M_t$.
\item Both sampling and predictive models are updated using $M_t$ through a single-epoch backpropagation step with $L_t(M_t,\theta)$ in accordance with the constraint defined in part \ref{ProbStatement}.
\end{itemize}

\section{Experiments}
The continual learning framework and attention-based ER is empirically evaluated through multivariate LT-TSF, assessing several data sampling strategies across benchmark datasets from the literature and real-world piezometric datasets. For this task, we chose as forecasting model the PatchMixer. The experiments were conducted on a server using a fixed random seed to ensure reproducibility. The hardware configuration included an Intel(R) Xeon(R) Gold 5220R CPU @ 2.20GHz and an NVIDIA RTX A6000 GPU.

\subsection{Datasets}
We use 12 multivariate time series datasets from diverse domains, including 3 standard benchmarks from the LT-TSF literature and 9 datasets related to groundwater level measurements in France:
\begin{itemize}
    \item ETTh2 and ETTm1: two subsets of the Electricity Transformer Temperature (ETT) benchmark \cite{Wu2021}, collected from a power substation. ETTh2 is recorded at an hourly resolution, while ETTm1 has a 15-minute resolution. Both include 7 variables (\textit{eg}. load, temperature, etc) spanning over two years.
    \item Weather: meteorological data collected across 21 U.S \cite{Wu2021}. cities, including variables such as temperature, pressure and wind speed, over a one-year period.
    \item Piezo: groundwater level datasets collected from piezometric sensors in France\footnote{Datasets available at "\href{https://hubeau.eaufrance.fr/}{hubeau.eaufrance}", with access to over 1,500+ piezometers.}. It includes 6 variables such as groundwater height and depth, along with related meteorological information (\textit{eg}. rainfall, temperature, etc). We choose 9 different samples from the same aquifer system with different behaviors to illustrate the model.
\end{itemize}

\begin{figure}[!ht]
\centering
\begin{subfigure}{0.48\linewidth}
    \centering
    \includegraphics[width=\linewidth]{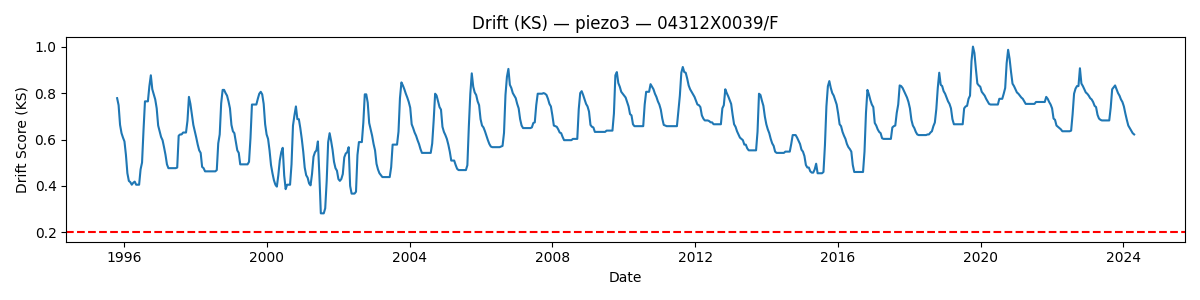}
    \caption{Drift Piezo 3}
\end{subfigure}
\hfill
\begin{subfigure}{0.48\linewidth}
    \centering
    \includegraphics[width=\linewidth]{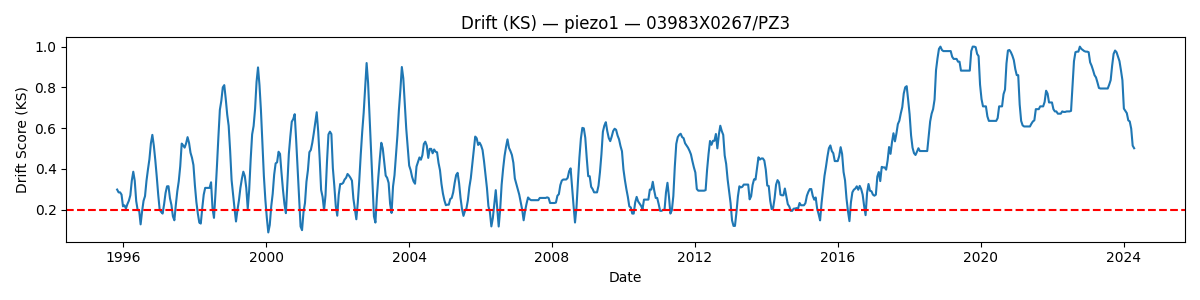}
    \caption{Drift Piezo 1}
\end{subfigure}
\caption{Drift analysis based on distributional changes measured with a sliding-window Kolmogorov–Smirnov test.}
\label{fig:piezo_drift}
\end{figure}
Some of the piezometric sensors in our study exhibit very slow temporal drift, which reduces the potential benefit of continual learning over static approaches. This is illustrated in Figure \ref{fig:piezo_drift} for piezo 3, with the distributional drift estimated using a sliding-window Kolmogorov–Smirnov test. These time series mainly show seasonal variability without any significant long-term drifts, and their spatial behavior remains stable and largely redundant. Consequently, continual learning should offer only little added value compared with conventional models when sufficient training data are available. In contrast, piezo 1 in Figure \ref{fig:piezo_drift} exhibits a pronounced and continuous drift, including a sudden shift at one point captured by the Kolmogorov–Smirnov test highlighting the non-stationary behavior of its underlying data.
\begin{figure}[!ht]
\centering
\begin{subfigure}{0.48\linewidth}
    \centering
    \includegraphics[width=\linewidth]{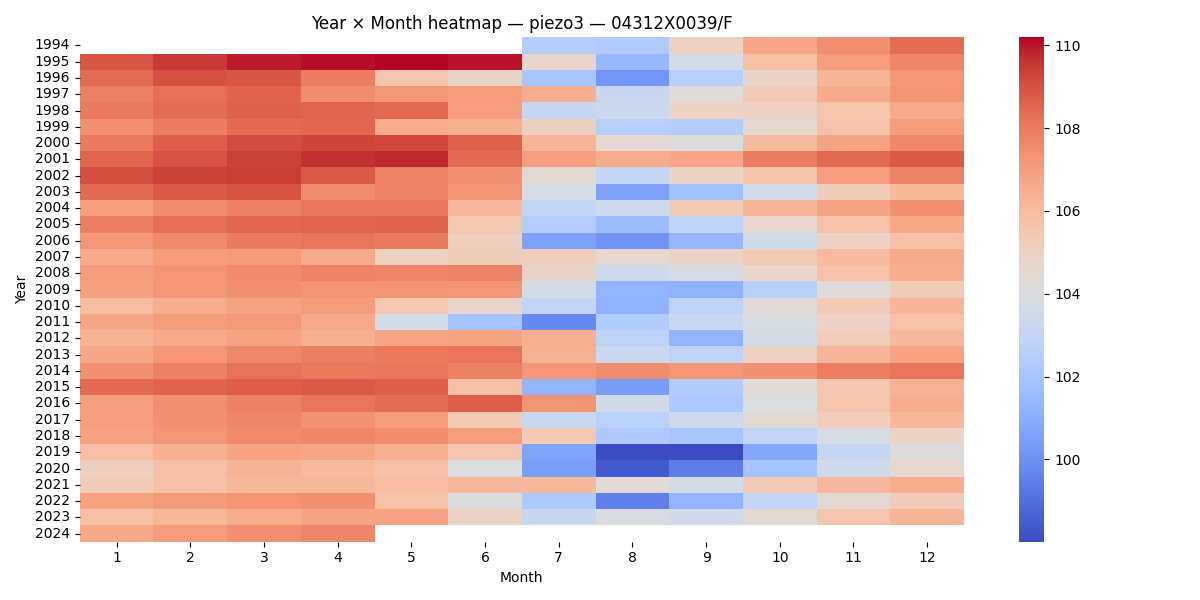}
    \caption{Heatmap Piezo 3.}
\end{subfigure}
\hfill
\begin{subfigure}{0.48\linewidth}
    \centering
    \includegraphics[width=\linewidth]{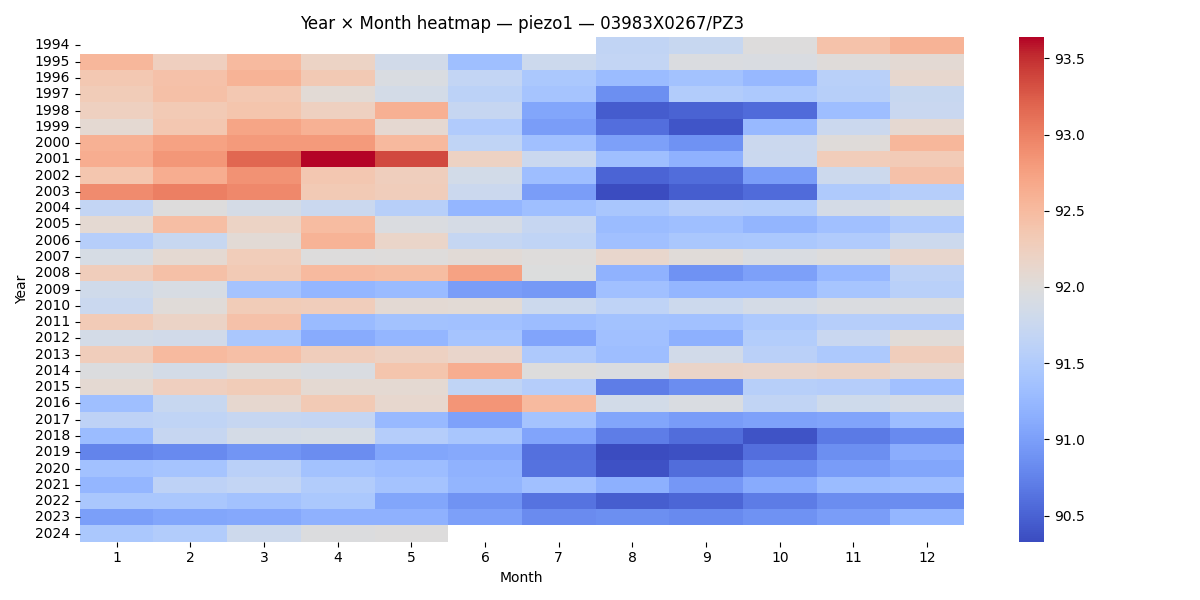}
    \caption{Heatmap Piezo 1.}
\end{subfigure}
\caption{Heatmap analysis on groundwater level change over time.}
\label{fig:piezo_heatmap}
\end{figure}
The piezometers heatmaps in Figure \ref{fig:piezo_heatmap} illustrate the groundwater level evolution from month-to-month (row wise) and year-to-year (column wise). High level recharge are in red whereas low levels are in blue. Piezo 1 exhibits a pronounced changing pattern, after several years making it a relevant case for continual learning due to its temporal variability. In contrast, Piezo 3 displays consistent and stable seasonal behavior, providing limited benefit for continual adaptation.

\subsection{Experimental Setup}
We consider a realistic scenario where the model is initially pretrained on a training subset and then continuously learned with batches that arrive sequentially during the testing phase:
\begin{itemize}
    \item Pretraining: The model is trained on 70\% of each dataset, following the PatchMixer reference setup.
    \item Application: The remaining 30\% is split into 10\% for validation and 20\% for testing, with metrics computed on the testing subset.
    \item Continual Adaptation: After one full epoch on the training set to initialize forecasting model and the MHAM model, the full system is tested and adapted sequentially on the testing subset, simulating a realistic deployment scenario with continual learning.
    \item LT-TSF: Input length is set to \texttt{seq\_len=336} and output length to \texttt{pred\_len=720}, corresponding to $\sim$2years for daily datasets (Piezometric), $\sim$1month for hourly datasets (ETTh2), and $\sim$1 week for high-frequency datasets (ETTm1, Weather).
\end{itemize}
Datasets from the literature are originally preprocessed. For piezometric data, any missing days are filled via linear interpolation without much preprocessing. In a continual learning setting, adaptive mechanisms for detecting and correcting outliers would be required, but distinguishing abrupt drifts from true anomalies is highly challenging.

\noindent These experiments were evaluated across five different learning strategies:
\begin{itemize}
\item Batch Learning: baseline case using separate training and testing subsets without any model adaptation.
\item Joint Training: similar to the batch learning setup but with continuous updates on all seen data during the testing phase (need to store all data). This strategy could be considered as the best one in average at a given moment.
\item CL: continual learning setting where a fixed size memory buffer is updated according to a given policy and the model is updated only on this buffer. We consider three policies :
\begin{itemize}
\item Random: the buffer is updated using uniformly sampled observations for replay.
\item MinLoss / MaxLoss: a loss-based policy is used where each sample is evaluated by its instantaneous MAE/MSE. The buffer retains either the top-$k$ Max Loss samples (to target model weaknesses) or the top-$k$ Min Loss samples (to reinforce already learned patterns). Here, we report only the Min Loss results, which consistently outperformed Max Loss on Target metrics.
\item MHAM: our proposed attention-based replay mechanism, which employs a continuously learned attention module to select the most informative samples for replay. 
\end{itemize}
\end{itemize}

Reported values correspond to the best performance observed across batch sizes \texttt{BS $\in$ {8, 16, 32, 64, 128}} with \texttt{buffer\_size = 250}, \texttt{past\_replay\_ratio = 0.2} and \texttt{current\_replay\_ratio = 0.2} (best parameters identified through empirical evaluation). Batch size is particularly important for the Attention strategy, as it determines how relationships among elements within a batch are modeled. 

\subsection{Evaluation Metrics}
We use two standard metrics for time series forecasting: Mean Absolute Error (MAE) and Mean Squared Error (MSE). These are evaluated at two levels:
\begin{itemize}
    \item Target performance: classical predictive accuracy on the test subset, reflecting the model’s performance on the most recent unseen data. The Target metric is computed after rescaling predictions to the original data distribution to better reflect realistic observed performance.
    \item Mean performance: accuracy on previously seen batches, measuring knowledge retention and resistance to catastrophic forgetting (Not feasible in real-world scenarios without storing all data).
\end{itemize}

For each metric, the best performances are highlighted when they fall within approximately 0.005 of the lowest observed value. Across metrics, the best scores are shown in bold for the Target evaluation and are underlined for the Mean evaluation.

\subsection{Results}
Tables \ref{tab:comparison1} and \ref{tab:comparison2} summarize the MAE \footnote{Similar results can be observed with MSE (see supplementary materials for more)} forecasting results for datasets exhibiting minimal or moderate drift (including literature benchmarks and several piezometric series). 

\begin{table}[!ht]
\centering
\begin{tabular}{ll|ccccc}
 & MAE & \makecell{ETTh2\\(BS=128)}& \makecell{WEATHER\\(BS=64)}& \makecell{Piezo 2\\(BS=8)}& \makecell{Piezo 3\\(BS=16)}& \makecell{Piezo 7\\(BS=128)}\\
\hline
Batch Learning & Target & \textbf{4.500}& 15.440& 1.866 & 1.957 & 1.587 \\
 & Mean & \underline{0.336}& 0.363 & 0.430& 0.536 & \underline{0.402}\\
Joint Training & Target & 4.549 & 14.600& \textbf{1.804}& \textbf{1.939}& \textbf{1.539}\\
 & Mean & \underline{0.335}& \underline{0.354}& \underline{0.412}& \underline{0.526}& 0.408 \\
CL Random & Target & 4.538 & 14.474 & 1.855 & 1.958 & 1.564 \\
 & Mean & \underline{0.333}& \underline{0.353}& 0.419 & 0.532 & \underline{0.396}\\
CL MinLoss & Target & 4.546 & \textbf{14.438}& 1.840& 1.977 & 1.542 \\
 & Mean & \underline{0.331}& \underline{0.356}& 0.418 & 0.584 & 0.413 \\
CL Attention & Target & 4.562 & 14.801 & 1.862 & 1.952& 1.574 \\
 & Mean & \underline{0.334}& \underline{0.358}& 0.419 & 0.535 & \underline{0.396}\\
\end{tabular}
\caption{MAE comparison of different Experience Replay strategies in case of no/few drift dataset.}
\label{tab:comparison1}
\end{table}
 
\begin{table}[!ht]
\centering
\resizebox{\textwidth}{!}{%
\begin{tabular}{ll|ccccccc}
 & MAE & \makecell{ETTm1\\(BS=128)}& \makecell{Piezo 1\\(BS=64)}& \makecell{Piezo 4\\(BS=128)}& \makecell{Piezo 5\\(BS=64)}& \makecell{Piezo 6\\(BS=32)}& \makecell{Piezo 8\\(BS=8)}& \makecell{Piezo 9\\(BS=16)}\\
\hline
Batch Learning & Target & 1.805 & 1.643 & \textbf{1.499}& 1.627 & 1.662 & 1.631 & \textbf{1.659}\\
 & Mean & 0.379 & \underline{0.610}& 0.622 & 0.515 & 0.596 & 0.400& 0.481 \\
Joint Training & Target & 1.772 & \textbf{1.631}& \textbf{1.502}& \textbf{1.562}& \textbf{1.634}& \textbf{1.599}& \textbf{1.661}\\
 & Mean & \underline{0.370}& \underline{0.607}& \underline{0.613}& \underline{0.505}& \underline{0.582}& \underline{0.389}& \underline{0.470}\\
CL Random & Target & \textbf{1.759}& \textbf{1.636}& 1.510& \textbf{1.565}& 1.639 & 1.604 & 1.667 \\
 & Mean & \underline{0.371}& \underline{0.607}& \underline{0.613}& \underline{0.507}& \underline{0.583}& \underline{0.392}& \underline{0.470}\\
CL MinLoss & Target & 1.780& 1.644 & \textbf{1.505}& 1.613 & 1.658 & 1.615 & 1.675 \\
 & Mean & \underline{0.373}& \underline{0.608}& \underline{0.617}& \underline{0.507}& 0.588 & \underline{0.392}& \underline{0.473}\\
CL Attention & Target & \textbf{1.758}& \textbf{1.633}& \textbf{1.501}& \textbf{1.568}& \textbf{1.629}& \textbf{1.601}& \textbf{1.665}\\
 & Mean & \underline{0.373}& \underline{0.607}& \underline{0.614}& \underline{0.509}& \underline{0.586}& \underline{0.392}& \underline{0.473}\\

\end{tabular}
}
\caption{MAE comparison of different Experience Replay strategies in case of visible drift dataset.}
\label{tab:comparison2}
\end{table}
\vspace*{-1cm}

\subsection{Analysis}
The results presented in Tables \ref{tab:comparison1} and \ref{tab:comparison2} highlight the respective strengths of different learning strategies under both stationary and non-stationary conditions. The batch learning approach maintains good performance on certain datasets (ETTh2, Piezo 3, 4 and 9), where the data distribution remains mostly constant over time. On datasets with limited drift (ETTh2, WEATHER and some piezo sensors), the offline PatchMixer is already showing strong predictive capability. 

Joint training, which uses the full dataset at each step, represents the performance upper bound. It achieves the best Mean and Target results for several benchmark thanks to complete data access, but at the cost of high computational demands. CL approach are showing close performance (especially for Mean metric) at a lower computational cost.

In case of datasets with visible drift (ETTm1 and other piezo), the proposed CL method (Attention Sampling) is always as good or better than JT while being much less time consuming which is more visible in Table.\ref{tab:comparison3} and Table.\ref{tab:comparison4}.  This show that the attention mechanism effectively prioritizes replay samples, preserving forecasting accuracy while minimizing scaling-related distortions and forgetting at a lower cost than a complete retraining every time a new batch of data is available.

\begingroup
\setlength{\tabcolsep}{3pt}
\begin{table}[!ht]
\centering

\begin{tabular}{l|ccccccc|c}
Target & ETTm1 & Piezo1 & Piezo4 & Piezo5 & Piezo6 & Piezo8 & Piezo9 & Average\\
\hline
Batch Learning & 2.6\% & 0.6\% & -0.1\% & 3.6\% & 2.0\% & 1.8\% & -0.4\% & 1.4\% \\
Joint Training & 0.8\% & -0.1\% & 0.1\% & -0.4\% & 0.3\% & -0.1\% & -0.2\% & 0.0\% \\
CL Random & 0.1\% & 0.2\% & 0.6\% & -0.2\% & 0.6\% & 0.2\% & 0.1\% & 0.2\% \\
CL MinLoss & 1.2\% & 0.7\% & 0.3\% & 2.8\% & 1.7\% & 0.9\% & 0.6\% & 1.2\% \\
\hline
Average & 1.2\% & 0.4\% & 0.2\% & 1.5\% & 1.2\% & 0.7\% & 0.0\% & 0.7\% \\

\end{tabular}
\caption{MAE gains of CL Attention compared to other configurations (\%).}
\label{tab:comparison3}
\end{table}
\endgroup
The random continual learning (CL) approach is particularly appealing in the no-drift setting. While attention-based mechanisms enable more effective temporal adaptation, the uniformity of random sampling offers consistently strong average performance and allows model updates at a very low computational cost. Random replay remains a common strategy in the literature and its resilience is further confirmed by our results too. Several improvement remain possible, particularly regarding buffer-related parameters. Parameters such as $k^{past}$ and $k^{curr}$ were selected to provide strong overall performance but could be tuned specifically for each application scenario. We currently use a training split corresponding to 70\% of the dataset, following the configuration of the PatchMixer repository. This assumes that a substantial amount of data is already available, thereby limiting the performance gains that continual adaptation can bring over standard batch learning.

\begingroup
\setlength{\tabcolsep}{5pt}
\begin{table}[!ht]
\centering
\begin{tabular}{lccccc}
\hline
 & BL & JT & Random & MinLoss & Attention \\
\hline
 Average runtime (min)& 0.71  & 37.0  & 3.43  & 8.86  & 5.0\\
\hline
\end{tabular}
\caption{Average update runtime across all piezo datasets.}
\label{tab:comparison4}
\end{table}
\endgroup
Table \ref{tab:comparison4} reports the average runtime of each method, including the cost of buffer management when used, on piezometric datasets. JT offers near-optimal accuracy but is about seven times slower than Attention, making it unscalable for CL with growing, high-frequency data. Attention requires only $\sim$1.6 minutes more than Random while improving Target error, yielding a strong accuracy and efficiency trade-off. With a buffer of 1,500 for datasets of ~10,000 points, only 15\% of the data is stored, while achieving performance close to full JT.

\section{Conclusion}
Real-world data such as the piezometric time series typically exhibit distributional shifts, regime changes and long-term drifts driven by dynamic environment. These characteristics introduce non-stationarity that challenges conventional forecasting models trained under the assumption of stable data. Standard batch learning models remain competitive when no major drift is observed. Nevertheless, in operational contexts, abrupt or unforeseen shifts may arise at any time. In such conditions, adaptive learning strategies become a critical requirement for continual data production while avoiding excessive retention.  

In this work, we addressed the challenge of adapting time series forecasting models to non-stationary environments using continual learning principles. We proposed a flexible framework to extend static models commonly used in the literature, such as PatchMixer, by incorporating an Experience Replay strategy guided by Attention mechanisms. This approach allows the model to adapt dynamically to new contexts while preserving prior knowledge, effectively mitigating catastrophic forgetting.

Extensive experiments on benchmark datasets (ETTm1, ETTh2, WEATHER) and real-world piezometric data demonstrate that the attention-based continual learning strategy achieves performance comparable to, and in many cases surpassing, traditional Joint Training approaches. It performs particularly well on non-stationary datasets, maintaining low prediction errors and stable long-term knowledge retention. Although Joint Training can perform well in static scenarios, its reliance on full data access and costly retraining limits its applicability in dynamic environments. In contrast, continual learning strategies (Random, MinLoss, Attention) provide competitive or superior performance while drastically reducing computational and storage requirements.

The proposed framework is lightweight and modular, requiring no modification of the underlying base forecasting model and relying on a dynamically updated buffer that integrates both historical and newly observed data. This data are carefully selected thanks to an attention sampling module continuously updated.
This design ensures stability during adaptation while maintaining low computational and energy costs, making it suitable for real-world deployments where efficiency is critical. Unlike Joint Training, which retrains regularly the model on all past and new data, continual learning provides a scalable and environmentally efficient alternative to classical retraining-based strategies. 

Overall, our study highlights the strong potential of the proposed continual learning strategy for robust, efficient adaptation of time series forecasting models in dynamic environments. Future work will explore complementary mechanisms such as structural growth and knowledge distillation, enabling models to adjust their capacity over time while controlling complexity through pruning. Combined with memory-based replay, these directions may further enhance performance, scalability, and sustainability.

%
%
%
\bibliographystyle{unsrt}
\bibliography{bibli}

\end{document}